\documentclass{article}

\usepackage{amsmath,amsfonts,amssymb,times,graphicx,natbib,algorithm,algorithmic}

\usepackage[accepted]{whi2020}

\icmltitlerunning{Pattern-Guided Integrated Gradients}
\usepackage{url}
\usepackage{xcolor}
\usepackage{filecontents}
\usepackage{pgfplots}
\usepackage{tikz}
\usetikzlibrary{arrows,backgrounds}
\usepgflibrary{shapes.multipart}

\usepackage{subfigure}

\begin{document}

\twocolumn[
\icmltitle{Pattern-Guided Integrated Gradients}


\icmlsetsymbol{equal}{*}

\begin{icmlauthorlist}
\icmlauthor{Robert Schwarzenberg*}{dfki}
\icmlauthor{Steffen Castle*}{dfki}
\end{icmlauthorlist}

\icmlaffiliation{dfki}{German Research Center for Artificial Intelligence (DFKI)}

\icmlcorrespondingauthor{Robert Schwarzenberg}{robert.schwarzenberg@dfki.de}

\icmlkeywords{interpretability, transparency}

\vskip 0.3in
]


\printAffiliationsAndNotice{\icmlEqualContribution}

\begin{abstract}
    Integrated Gradients (\textsc{IG}) and PatternAttribution (\textsc{PA}) are two established explainability methods for neural networks. Both methods are theoretically well-founded. However, they were designed to overcome different challenges. In this work, we combine the two methods into a new method, \linebreak \textit{Pattern-Guided Integrated Gradients} (\textsc{PGIG}). \textsc{PGIG} inherits important properties from both parent methods and passes stress tests that the originals fail. In addition, we benchmark \textsc{PGIG} against nine alternative explainability approaches (including its parent methods) in a large-scale image degradation experiment and find that it outperforms all of them. 
\end{abstract}

\section{Introduction}
Integrated Gradients \cite{sundararajan2017axiomatic} is a gradient-based explainability method with a sound theoretical motivation that has also performed well experimentally \cite{ancona2017towards}. The authors of the PatternAttribution method \cite{kindermans2017learning}, however, compellingly argue that IG does not give enough importance to the class signals in the input data (as do several other gradient-based explainability methods). They show that model weights function to cancel the distractor (noise) in the data and thus are more informative about the distractor than the signal. The weights even tend to direct the gradient (and hence attributions of gradient-based explainability methods) away from the signal. To direct the weights (and thus attributions) towards the signal, PA modifies them with informative directions -- called patterns --- that it learns from data. 

\begin{figure}[!htb]
\centering
\subfigure[]{
\includegraphics[scale=.094]{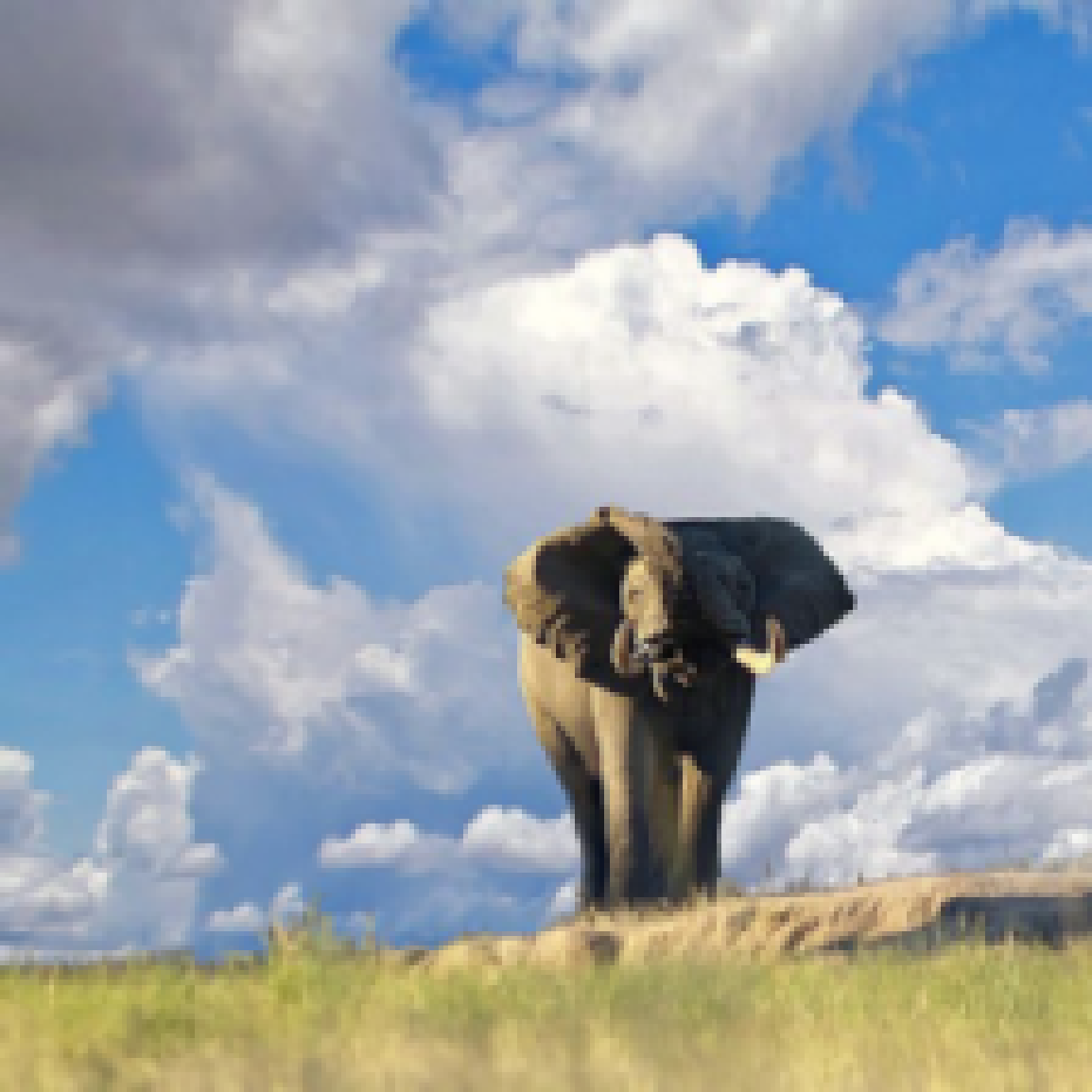}
}
\subfigure[]{
\includegraphics[scale=.094]{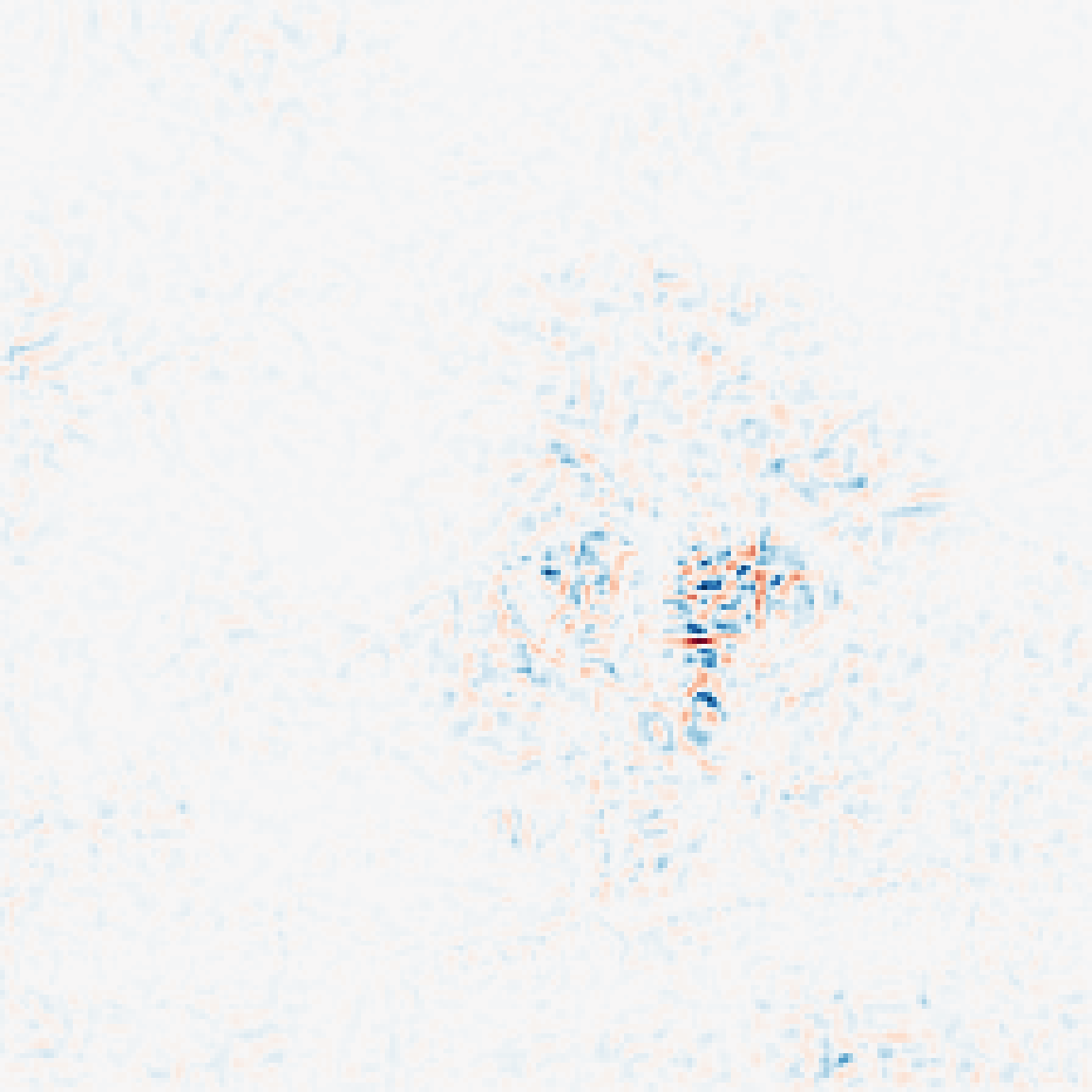}
}
\subfigure[]{
\includegraphics[scale=.094]{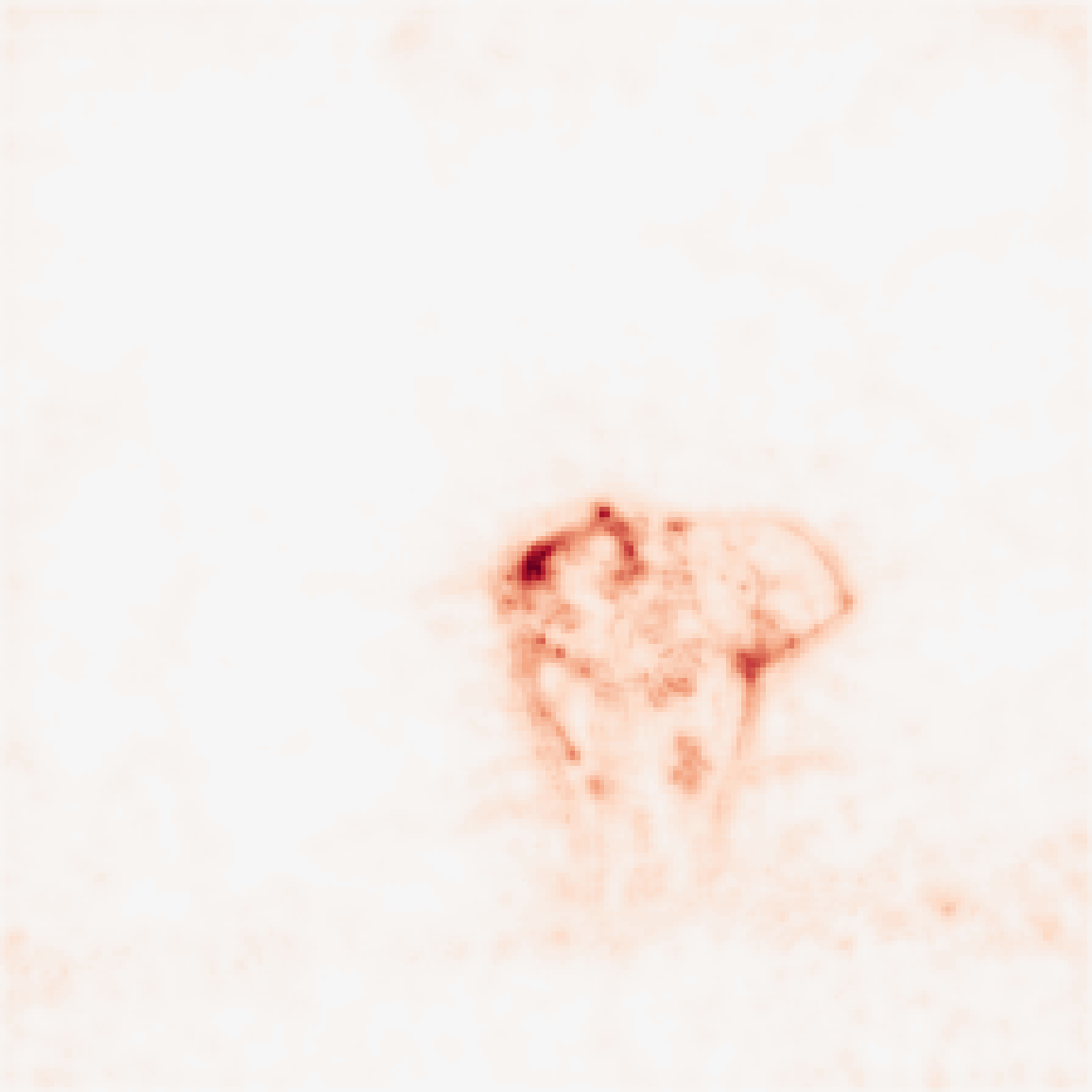}
}
\subfigure[]{
\includegraphics[scale=.094]{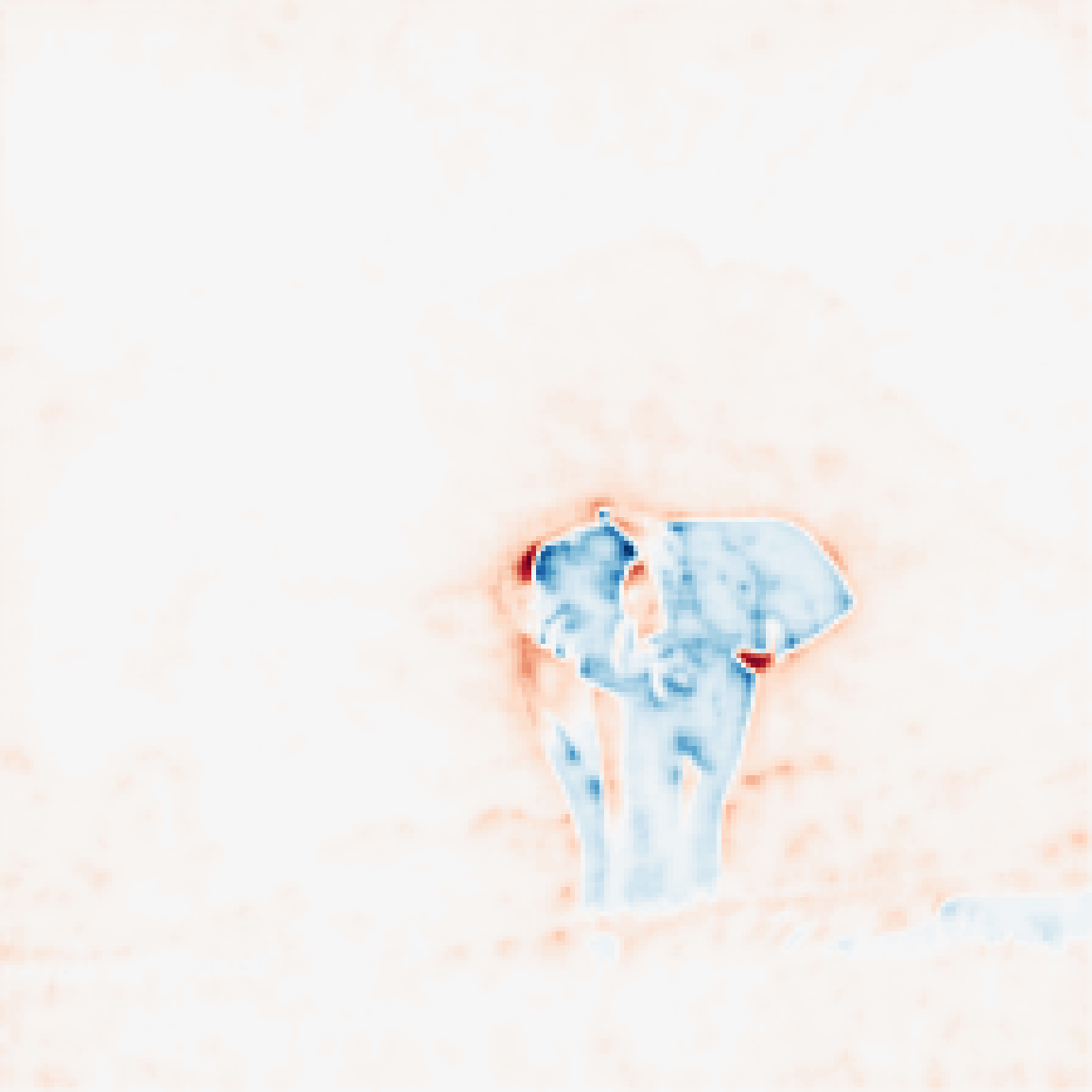}
}
\caption{Saliency maps (red: positive, blue: negative) for an input (a) according to IG (b), PA (c), and PGIG (d), explaining the correct (\texttt{African Elephant}) VGG-16 classification of (a).} 
\label{fig:heatmaps}
\end{figure}

We show that PA, in turn, suffers from  problems that IG has overcome. In particular, it suffers from the saturation problem which occurs at function plateaus (cf. vanishing gradient), resulting in zero-attributions for input features that contributed to non-zero output activations.

In response, we propose a hybrid approach, Pattern-Guided Integrated Gradients, that combines the strengths of both methods. We demonstrate that \textsc{PGIG} passes controlled stress tests that both parent methods fail. Furthermore, we find that it outperforms its parent methods, as well as seven other prominent explainability methods, in a large-scale image degradation experiment. The new method can be implemented quickly, in particular when Integrated Gradients and PatternAttribution are already part of the code base, as is the case in several explainability frameworks \cite{visualattr2018, alber2019innvestigate}.
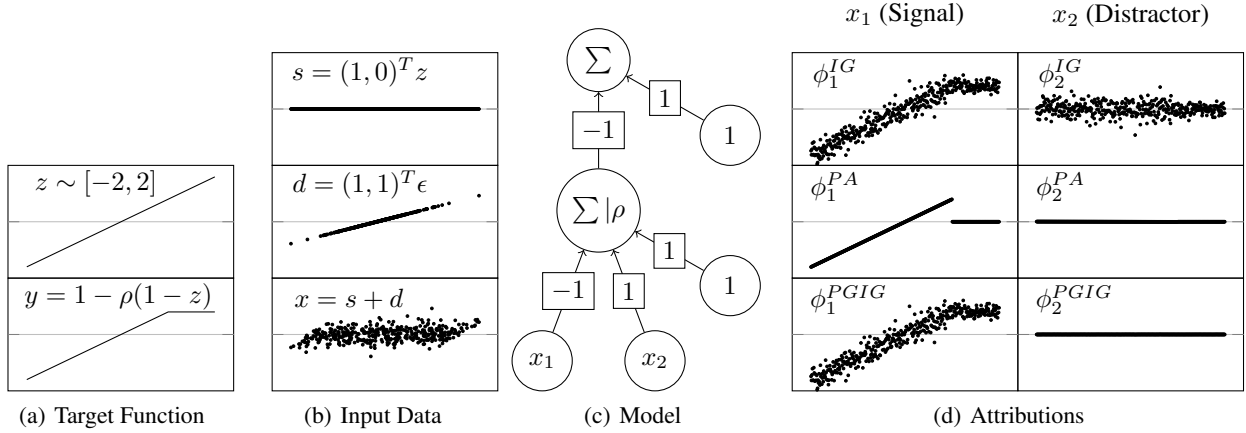
\begin{figure*}[!htb]
\centering

\subfigure[Target Function]{
\begin{tikzpicture}[
    /pgfplots/scale only axis,
    /pgfplots/width=3cm,
]
\begin{axis}[
    name=z,
    anchor=south east,
    height=1.5cm,
    xtick=\empty,
     extra y ticks       = 0,
      extra y tick labels = ,
      extra y tick style  = { grid = major},
      ymin=-2.5, ymax=2.5,
    ytick=\empty,
    y label style={at={(axis description cs:0.8,.64)}, rotate=-90, anchor=south},
    ylabel={$z \sim [-2,2]$},
     x label style={at={(axis description cs:0.5,+1.5)},anchor=south},
]
\addplot[] 
table [col sep=comma] {csvs/z.csv};
\end{axis}
\begin{axis}[
    name=y,
    anchor=north west,
    at=(z.south west),
    height=1.5cm,
    xtick=\empty,
     extra y ticks       = 0,
      extra y tick labels = ,
      extra y tick style  = { grid = major},
      ymin=-2.5, ymax=2.5,
    ytick=\empty,
    y label style={at={(axis description cs:0.9,.64)}, rotate=-90, anchor=south},
    ylabel={$y = 1-\rho(1-z)$},
     x label style={at={(axis description cs:0.5,+1.5)},anchor=south},
]
\addplot[] 
table [col sep=comma] {csvs/y.csv};
\end{axis}
\end{tikzpicture}
} 
\subfigure[Input Data]{
\begin{tikzpicture}[
    /pgfplots/scale only axis,
    /pgfplots/width=3cm,
]
\begin{axis}[
    name=s,
    anchor=south east,
    height=1.5cm,
    xtick=\empty,
     extra y ticks       = 0,
      extra y tick labels = ,
      extra y tick style  = { grid = major, color=red },
      ymin=-2.5, ymax=2.5,
    ytick=\empty,
    y label style={at={(axis description cs:0.8,.62)}, rotate=-90, anchor=south},
    ylabel={$s = (1,0)^{T}z$},
     x label style={at={(axis description cs:0.5,+1.5)},anchor=south},
]
\addplot[only marks, black, mark size=.5pt] 
table [col sep=comma] {csvs/signal.csv};
\end{axis}
\begin{axis}[
    name=d,
   anchor=north west,
    at=(s.south west),
    height=1.5cm,
    xtick=\empty,
     extra y ticks       = 0,
      extra y tick labels = ,
      extra y tick style  = { grid = major, color=red },
      ymin=-2.5, ymax=2.5,
    ytick=\empty, 
     ylabel={$d = (1,1)^{T}\epsilon$},
     y label style={at={(axis description cs:0.8,.62)}, rotate=-90, anchor=south},
     x label style={at={(axis description cs:0.5,+1.5)},anchor=south},
]
\addplot[only marks, black, mark size=.5pt] 
table [col sep=comma] {csvs/distractor.csv};
\end{axis}

\begin{axis}[
    name=d,
   anchor=north west,
    at=(d.south west),
    height=1.5cm,
    xtick=\empty,
     extra y ticks       = 0,
      extra y tick labels = ,
      extra y tick style  = { grid = major, color=red },
      ymin=-2.5, ymax=2.5,
    ytick=\empty, 
     ylabel={$x = s + d$},
     y label style={at={(axis description cs:0.75,.64)}, rotate=-90, anchor=south},
]
\addplot[only marks, black, mark size=.5pt] 
table [col sep=comma] {csvs/X.csv};
\end{axis}

\end{tikzpicture}}
\subfigure[Model]{
    \begin{tikzpicture}
    	\tikzstyle{place}=[circle, draw=black, minimum size = 8mm]
    	    \draw node at (0, 0) [place] (first_1) {$x_1$};
    	    \draw node at (1.5, 0) [place] (first_2) {$x_2$};
    	    \draw node at (2.5, 1) [place] (first_b) {$1$};
    	    
    	    \draw node at (.75, 2) [place] (second_1) {$\sum | \rho$};
    	    \draw [->] (first_1) -- (second_1) node[draw, midway, fill=white]{$-1$};
    	   \draw [->] (first_2) -- (second_1) node[draw, fill=white, midway]{$1$};
    	   \draw [->] (first_b) -- (second_1) node[draw, midway, fill=white]{$1$};
    	   
    	   \draw node at (.75, 4) [place] (third_1) {$\sum$};
    	   \draw node at (2.5, 3) [place] (second_b) {$1$};
    	   \draw [->] (second_1) -- (third_1) node[draw, midway, fill=white]{$-1$};
    	   \draw [->] (second_b) -- (third_1) node[draw, midway, fill=white]{$1$};
    	
    	
    		
    	
    	
    \end{tikzpicture}}
\subfigure[Attributions]{
\begin{tikzpicture}[
    /pgfplots/scale only axis,
    /pgfplots/width=3cm,
]
\begin{axis}[
    name=intgradx2,
    height=1.5cm,
    xtick=\empty,
     extra y ticks       = 0,
      extra y tick labels = ,
      extra y tick style  = { grid = major },
      ymin=-2.5, ymax=2.5,
    ytick=\empty, x label style={at={(axis description cs:0.5,+1.5)},anchor=south},
    xlabel={$x_{2}$ (Distractor)}, 
    y label style={at={(axis description cs:0.60,.6)}, rotate=-90, anchor=south},
    ylabel={$\phi^{IG}_{2}$}
]
\addplot[only marks, black, mark size=.5pt] 
table [col sep=comma] {csvs/IG_distractor.csv};
\end{axis}
\begin{axis}[
    name=intgradx1,
    anchor=south east,
    at=(intgradx2.south west),
    height=1.5cm,
    xtick=\empty,
     extra y ticks       = 0,
      extra y tick labels = ,
      extra y tick style  = { grid = major, color=red },
      ymin=-2.5, ymax=2.5,
      ytick=\empty, 
       x label style={at={(axis description cs:0.5,+1.5)},anchor=south},
    xlabel={$x_{1}$ (Signal)},
    y label style={at={(axis description cs:0.60,.6)}, rotate=-90, anchor=south},
    ylabel={$\phi^{IG}_{1}$}
]
\addplot[only marks, black, mark size=.5pt] 
table [col sep=comma] {csvs/IG_signal.csv};
\end{axis}

\begin{axis}[
    name=patattx2,
    anchor=north west,
    at=(intgradx2.south west),
    height=1.5cm,
    xtick=\empty,
     extra y ticks       = 0,
      extra y tick labels = ,
      extra y tick style  = { grid = major },
      ymin=-2.5, ymax=2.5,
    ytick=\empty, 
      y label style={at={(axis description cs:1.55,.5)}, anchor=south},
    y label style={at={(axis description cs:0.60,.6)}, rotate=-90, anchor=south},
    ylabel={$\phi^{PA}_{2}$}
]
\addplot[only marks, black, mark size=.5pt] 
table [col sep=comma] {csvs/PA_distractor.csv};
\end{axis}
\begin{axis}[
    name=patattx1,
    anchor=south east,
    at=(patattx2.south west),
    height=1.5cm,
    xtick=\empty,
     extra y ticks       = 0,
      extra y tick labels = ,
      extra y tick style  = { grid = major, color=red },
      ymin=-2.5, ymax=2.5,
    ytick=\empty,
    y label style={at={(axis description cs:0.60,.6)}, rotate=-90, anchor=south},
    ylabel={$\phi^{PA}_{1}$}
]
\addplot[only marks, black, mark size=.5pt] 
table [col sep=comma] {csvs/PA_signal.csv};
\end{axis}

\begin{axis}[
    name=pgigx2,
    anchor=north west,
    at=(patattx2.south west),
    height=1.5cm,
    xtick=\empty,
     extra y ticks       = 0,
      extra y tick labels = ,
      extra y tick style  = { grid = major },
      ymin=-2.5, ymax=2.5,
    ytick=\empty,
    y label style={at={(axis description cs:0.66,.6)}, rotate=-90, anchor=south},
    ylabel={$\phi^{PGIG}_{2}$}
]
\addplot[only marks, black, mark size=.5pt] 
table [col sep=comma] {csvs/PGIG_distractor.csv};
\end{axis}
\begin{axis}[
    name=pgigx1,
    anchor=south east,
    at=(pgigx2.south west),
    height=1.5cm,
    xtick=\empty,
     extra y ticks       = 0,
      extra y tick labels = ,
      extra y tick style  = { grid = major, color=red },
      ymin=-2.5, ymax=2.5,
    ytick=\empty,
    y label style={at={(axis description cs:0.66,.6)}, rotate=-90, anchor=south},
    ylabel={$\phi^{PGIG}_{1}$}
]
\addplot[only marks, black, mark size=.5pt] 
table [col sep=comma] {csvs/PGIG_signal.csv};
\end{axis}
\end{tikzpicture}
}
\caption{Stress tests for attribution methods, involving a plateau and noise, combining challenges from \citet{sundararajan2017axiomatic} and \citet{kindermans2017learning}. \textbf{a} shows the target function ($\rho$ is shorthand for ReLU, horizontal line located at zero). \textbf{b} visualizes synthetic input training data $x$ (\textbf{b}, bottom row). The data is composed of a signal $s$ (\textbf{b}, top row) that introduces information about the target to the first dimension of $x$ and a distractor $d$ (\textbf{b}, middle row) that introduces noise to both dimensions of $x$. Note that the second dimension of $x$ does not carry information about $y$, just noise. \textbf{c} depicts a model (weights in rectangles) which has learned the mapping $x \rightarrow y$ by effectively using the signal in $x_{1}$ and cancelling the noise from the distractor in $x_{2}$. \textbf{d} shows attribution scores for the input features according to Integrated Gradients (\textbf{d}, top row), PatternAttribution (\textbf{d}, middle row) and the proposed Pattern-Guided Integrated Gradients (\textbf{d}, bottom row). Only the proposed method attributes importance to the signal in $x_{1}$ at the \textit{non-zero} plateau (despite a zero-gradient) and bypasses the noise from the distractor in $x_{2}$.}
\label{fig:stresstest}
\end{figure*}
\section{Prerequisites}
\label{sec:prerequisites}
In this section, we briefly discuss Integrated Gradients and PatternAttribution to introduce notation as well as important concepts and properties of the two attribution methods. We consider an attribution method a function $\phi_{f,i}(x): \mathbb{R}^{D} \rightarrow \mathbb{R}$ that maps each input feature $i\in\{1\dots D\}$ in $x \in \mathbb{R}^{D}$ to a real number that signifies the importance of input $i$ to the output of model $f: \mathbb{R}^{D} \rightarrow \mathbb{R}$.\footnote{For simplicity, without the loss of generality, we assume one-dimensional outputs throughout this paper.}

\subsection{Integrated Gradients}
\label{secsub:integrated-gradients}

The attributions provided by Integrated Gradients are a summation of the gradient attribution maps at values from the straight-line path between a baseline $\bar{x}$ (a user-defined reference point) and the input, $x$. The formula is given by
\begin{equation}
\label{eq:ig}
\phi_{f,i}^{IG}(x) = \frac{x_{i} - \bar{x}_{i}}{m}\sum_{k=1}^{m} \frac{\partial f(\bar{x} + \frac{k}{m}(x - \bar{x}))}{\partial x_{i}}
\end{equation}

where $m$ is a hyperparameter, the number of equidistant steps along the path. As a path method, IG mitigates the aforementioned saturation problem, which we demonstrate below. Furthermore, the authors of Integrated Gradients cite two desirable properties for attribution methods: The first is referred to as \textit{Sensitivity}. An attribution method is \textit{sensitive} ``if  for  every input and baseline that differ in one feature but have different predictions then the differing feature should be given a non-zero attribution'' \cite{sundararajan2017axiomatic}.  The second property is called \textit{Implementation Invariance} and demands that two networks that are functionally equivalent -- regardless of implementation -- should always yield identical attribution maps.  IG is both sensitive and implementation invariant as $\lim m \rightarrow \infty$.

\subsection{PatternAttribution}
\label{secsub:patternattribution}
The authors of PatternAttribution  \cite{kindermans2017learning} criticize Integrated Gradients, among other gradient methods, for not discriminating signal and distractor in the input data. Their argument is based on the observation that a well-trained model cancels the distractor in the input -- that is, everything that it did not find to co-vary with the target. 
For example, assume that we want to model a simple linear relation $x \rightarrow y$, where $x = s + d$ is composed of a signal $s$ that carries all the information needed to predict the target $y$ (it co-varies with the target) and an additive distractor $d$. 
In case of a well-trained linear model, the model must have learned weights $w$ s.t.  $f(d)=w^{T}d =0$ and $f(x)=w^{T}x = w^{T}s =y$. 

Thus, the weights function as a filter that must always change direction with the distractor in order to stay orthogonal to it. A change in the signal, on the other hand, is accounted for by a change in magnitude of the weights. The authors conclude that gradient methods -- including IG -- that channel attributions along $w$ inherently direct it towards a direction that is determined by the distractor, not the signal. 

For PatternAttribution, prior to the backward pass, the weights $w$ of a linear or ReLU activated layer are replaced by $w \odot p$ where 
\begin{equation}
\label{eq:lincase}
p = \frac{\mathbb{E}_{+}[\mathbf{x}\mathbf{\hat{y}}]-\mathbb{E}_{+}[\mathbf{x}]\mathbb{E}[\mathbf{\hat{y}}]}{w^{T}\mathbb{E}_{+}[\mathbf{x}\mathbf{\hat{y}}]-w^{T}\mathbb{E}_{+}[\mathbf{x}]\mathbb{E}[\mathbf{\hat{y}}]}
\end{equation}
is a pattern computed over a batch of layer inputs and outputs $\mathbf{x}, \mathbf{\hat{y}}$. $\mathbb{E}_{+}[\cdot]$ denotes the expectation tensor over the positive regime of a ReLU activated layer, $\{x | w^{T}x > 0\}$. We can interpret Eq.~\ref{eq:lincase} as follows: Weights that primarily cancel the distractor are scaled down whereas weights that amplify or conserve the signal are preserved. This way, PatternAttribution directs a modified gradient
towards the signal. Subsequently, we denote a gradient backward call with the patterns in place as $\partial^{(p)}$. According to this notation, PatternAttribution becomes 
\begin{equation}
\phi^{\textsc{PA}}_{f,i}(x) = \frac{\partial^{(p)} f (x)}{ \partial^{(p)} x_{i}}
\end{equation}

\definecolor{random}{rgb}{1,.1,0}
\definecolor{valgrad}{rgb}{.9,.3,.1}
\definecolor{gradxinput}{rgb}{.8,.3,.2}
\definecolor{ig}{rgb}{.7,.3,.3}
\definecolor{vargrad}{rgb}{.6,.3,.4}
\definecolor{smoothgrad}{rgb}{.5,.3,.5}
\definecolor{smoothgrad-ig}{rgb}{.4,.3,.6}
\definecolor{expgrad}{rgb}{.3,.3,.7}
\definecolor{guidedbackprop}{rgb}{.2,.3,.8}
\definecolor{patternattribution}{rgb}{.1,.3,.9}
\definecolor{pgig}{rgb}{0,.1,1}

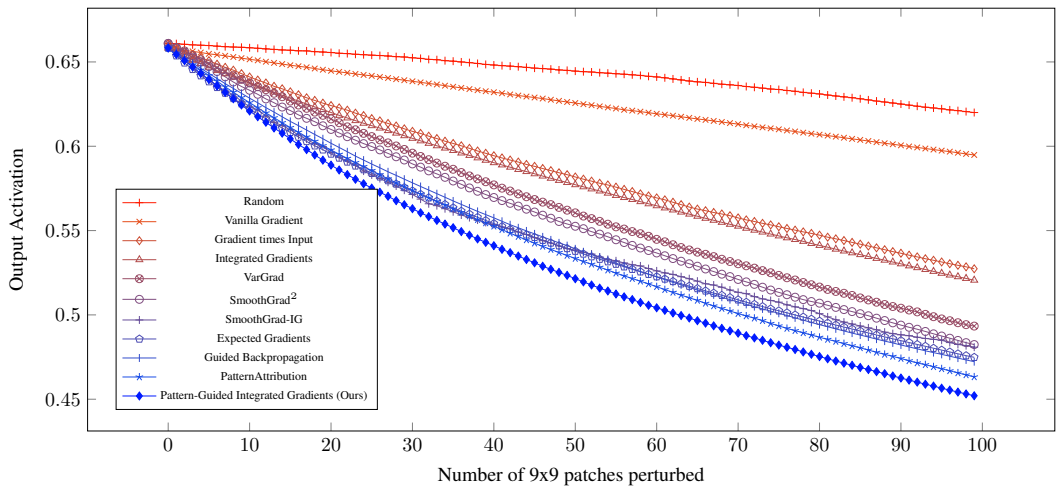
\begin{figure*}[!htb]
\centering
\begin{tikzpicture}[scale=.75, transform shape]
\begin{axis}[
scale only axis,
width=\textwidth,
height=7.5cm,
xlabel=Number of 9x9 patches perturbed,
ylabel=Output Activation,
legend style={at={(.3,0.05)},font=\tiny, anchor =south east,draw=black,fill=white,align=left},
]

\addplot [mark=+,color=random] table [x=patches, y=random, col sep=comma] {data.csv};
\addlegendentry{Random};

\addplot[mark=x, color=valgrad] table [x=patches, y=vanilla_grad, col sep=comma] {data.csv};
\addlegendentry{Vanilla Gradient};

\addplot [mark=diamond,color=gradxinput] table [x=patches, y=grad_x_input, col sep=comma] {data.csv};
\addlegendentry{Gradient times Input};

\addplot[mark=triangle, color=ig]  table [x=patches, y=integrate_grad, col sep=comma] {data.csv};
\addlegendentry{Integrated Gradients};

\addplot [mark=otimes,color=vargrad] table [x=patches, y=var_grad, col sep=comma] {data.csv};
\addlegendentry{VarGrad};

\addplot [mark=o,color=smoothgrad] table [x=patches, y=smooth_grad2, col sep=comma] {data.csv};
\addlegendentry{SmoothGrad$^{2}$};

\addplot [mark=+,color=smoothgrad-ig] table [x=patches, y=sg_ig, col sep=comma] {data.csv};
\addlegendentry{SmoothGrad-IG};

\addplot [mark=pentagon,color=expgrad] table [x=patches, y=exp_grad, col sep=comma] {data.csv};
\addlegendentry{Expected Gradients};

\addplot [mark=|,color=guidedbackprop] table [x=patches, y=guided_backprop, col sep=comma] {data.csv};
\addlegendentry{Guided Backpropagation};


\addplot[mark=star, color=patternattribution] table [x=patches, y=pattern_vanilla_grad, col sep=comma] {data.csv};
\addlegendentry{PatternAttribution};

\addplot[mark=diamond*, color=pgig]  table [x=patches, y=pattern_integrate_grad, col sep=comma] {data.csv};
\addlegendentry{Pattern-Guided Integrated Gradients (Ours)};

\end{axis}
\end{tikzpicture}
\caption{Image degradation experiment using VGG-16 on the val. split of ImageNet (steeper is better).}
\label{fig:exp-pert-ig}
\end{figure*}
\section{Stress Tests}

The authors of Integrated Gradients and the authors of PatternAttribution each present a different stress test to demonstrate the benefits of their method over alternative approaches. The former demonstrate that IG mitigates the saturation problem, the latter prove that PA is able to avoid noise.

We combine the two stress tests by defining a target function that involves a plateau and training a network to model the function with noisy input data. We demonstrate that each method fails the other's test: PA starves at the plateau and IG attributes importance to the noise in the input. We then combine the two methods into Pattern-Guided Integrated Gradients and demonstrate that the hybrid approach passes all tests. We illustrate this argument in Fig.~\ref{fig:stresstest}. 

\subsection{Target Function}  With IG, PA and PGIG we will later explain a neural network that models $y = 1-ReLU(1-z)$, for $z \in [z^{(1)} = -2, z^{(2)} = -1.99, \dots, z^{(N)} = 2]$, the target function, depicted in Fig.~\ref{fig:stresstest} (\textbf{a}). Please note that $y$ plateaus after $z>1$. This \textit{non-zero} plateau is the first stress test for the gradient-based attribution methods because the gradient becomes zero at the plateau but the attribution scores should not be zero. 

\subsection{Input Data} Let us now generate two-dimensional input training data, $x\in \mathbb{R}^{2}$ where $x = s + d$. As mentioned above, we want the signal $s$ to co-vary with the target. To generate such a signal, we scale $(1,0)^{T}$ with $z$, s.t. $\mathbf{s} = [(1,0)^{T}z^{(1)},(1,0)^{T}z^{(2)}, ...]$. The signal is visualized in Fig.~\ref{fig:stresstest} (\textbf{b}), top row. 

The distractor (Fig.~\ref{fig:stresstest} (\textbf{b}), middle row) is $d = (1,1)^{T}\epsilon$ where $\epsilon \sim \mathcal{N}(\mu, \sigma^{2})$, which we sample independently for each $s\in\mathbf{s}$. Note that while $s$ carries information about $z$ and $y$ in the first dimension, $d$ contains only noise, i.e. it does not contain information about the target. Because only $d$ is present in the second dimension, let us subsequently refer to the second dimension as the \textit{distractor dimension} and the first dimension as the \textit{signal dimension}. 

\subsection{Model} To produce the input given the target, the network must effectively cancel the distractor (below, we will construct such a network). This is the second stress test: If the model cancels the distractor, inputs pertaining only to the distractor should receive only zero attributions. In our case, this means that $x_{2}$ should receive only zero attributions, as it contains nothing but noise from the distractor. This is challenging since the respective inputs and weights might not be zero. 

Let us first consider a proxy model which learns $w$ such that $w^{T}x = z$. We can solve for $w$ analytically: Since $w$ needs to be orthogonal to the distractor base vector which is $(1,1)^{T}$, $w = (1,-1)^{T}$. If, for example, $s^{T} = (.5, 0)$ and $d^{T} = (.1, .1)$, then indeed $w^{T}(s + d) = .5 = z = w^{T}s$ and $w^{T}d = 0$. Thus, $w$ successfully cancels the distractor. 

Now, assume that $f^{(1)}$ and $f^{(2)}$ are two dense layers, with unit biases and parameters $w^{(1)}, w^{(2)}$, accepting two- and one- dimensional inputs, respectively. If we set $w^{(1)} = -w$ and $w
^{(2)} = (-1)$ then $y = f(x) = f^{(1)}(\rho(f^{(2)}(x)))$, where $\rho$ is shorthand for ReLU. This model is outlined in Fig.~\ref{fig:stresstest} (\textbf{c}).   

\subsection{Attributions} At this point, we have combined the two stress tests from \citet{sundararajan2017axiomatic} and \citet{kindermans2017learning} and apply IG and PA to receive attributions for $\hat{y}$, depicted in Fig~\ref{fig:stresstest} (\textbf{d}). 
For PA, we compute the patterns for $w^{(1)}$ and $w^{(2)}$ with Eq.~\ref{eq:lincase}, which yields $p^{(1)} \approx (-1, 0)^{T}$ and $p^{(2)} = (-1)$. The bias contributions are considered zero, as the bias does not co-vary with the target \cite{kindermans2017learning}. For PA, the backpropagation is started with $\hat{y}$, whereas for IG, the backpropagation is invoked starting with $1.0$.\footnote{For demonstration purposes, we violate the constraint for IG that output values are in the range $[0,1]$. In our case, this only scales the attributions and does not corrupt the method.}

IG (Fig.~\ref{fig:stresstest} (\textbf{d}), top row) follows the function in the signal dimension ($x_{1}$) including the plateau after $z>1$, which we consider desirable. It does, however, attribute a significant portion of importance to inputs from the distractor dimension, which only carries noise and is cancelled by the model. PA (Fig.~\ref{fig:stresstest} (\textbf{d}), middle row) successfully avoids the noise, i.e. its attribution scores in the distractor dimension are low, but it suffers from the saturation problem at the plateau in the signal dimension due to a zero-gradient. As a consequence, PA violates sensitivity. 
\section{Pattern-Guided Integrated Gradients}
 
 In response, we propose Pattern-Guided Integrated Gradients, given by
 \begin{equation}
 \label{eq:pgig}
     \phi_{f,i}^{PGIG}(x) = \frac{x_{i} - \bar{x}_{i}}{m}\sum_{k=1}^{m} \frac{\partial^{(p)} f(\bar{x} + \frac{k}{m}(x-\bar{x}))}{\partial^{(p)} x_{i}}
 \end{equation}
PGIG sums over the saliency maps returned by PA for inputs along the straight path between the baseline and the point of interest, $x$. 

\subsection{Properties}

Like IG, PGIG is a path method that mitigates the saturation problem and like PA, it considers informative directions and thus avoids the distractor. Its favorable attributions are visualized in the bottom row of Fig.~\ref{fig:stresstest} (\textbf{d}).

In the linear case, we can extract the pattern from Eq.~\ref{eq:pgig} and recover IG:
\begin{align*}
    \phi^{PGIG}_{f,i}(x) &= \frac{x_{i} - \bar{x}_{i}}{m}\sum_{k=1}^{m} \frac{\partial^{(p)} f(\dots)}{\partial^{(p)} x_{i}}\\ &= \frac{x_{i} - \bar{x}_{i}}{m}\sum_{k=1}^{m}p_{i}\frac{\partial f(\dots)}{\partial x_{i}}\\ &= p_{i}\phi_{f,i}^{IG}(x).
\end{align*}
PGIG scales the attribution scores of Integrated Gradients according to the class informativeness of input $i$. PGIG thus is sensitive to changes in all input dimensions $1 \leq i \leq D$ except for when $p_{i} = 0$. One interpretation of this is that PGIG is not sensitive to changes in pure distractor dimensions, such as $x_{2}$ in Fig.~\ref{fig:stresstest} (\textbf{c}), (\textbf{d}). This reasoning directly translates to intermediate and ReLU-activated layers. Regarding implementation invariance, we leave it to future work to prove or disprove this property for PGIG.
\section{Experiments}
\label{sec:pg-ig-experiments}
\citet{sundararajan2017axiomatic} motivate IG axiomatically but do not study their method empirically. \citet{kindermans2017learning} derive their method axiomatically as well, but they also conduct image degradation experiments, an established metric to estimate the quality of saliency methods. 

For the image degradation benchmark, a growing number of patches in the input images  are replaced with their mean channel values and the output activation\footnote{We explain networks after the final softmax activation.} (confidence) of the model is monitored. The order in which patches are perturbed is dictated by the accumulated saliencies of their pixels. The premise of this experiment is that the steeper the drop in confidence, the more accurately the attribution method has identified the most important features.

We, too, benchmark PGIG with the image degradation metric. Like \citet{kindermans2017learning}, we use a pre-trained VGG-16 model \cite{simonyan2014very} to generate saliency maps for the 50k images in the validation split of the ImageNet data set \cite{deng2009imagenet} that are then used to successively degrade the top 100 patches in descending order and aggregate confidence values. Images were cropped to 224x224 and normalized within $[-1,1]$. Our code base builds on the PyTorch Visual Attribution framework \cite{visualattr2018} from where we also received the patterns for VGG-16. Data and code are open source: \url{https://github.com/DFKI-NLP/pgig}.

We compare the patch ranking by PGIG against a random ordering of the patches as well as against rankings determined by nine other gradient-based explainability methods, which are organized into two classes.    

The first class contains gradient aggregation methods of which Integrated Gradients is a member. Vargrad \cite{adebayo2018sanity} and SmoothGrad \cite{smilkov2017smoothgrad} calculate the variance and mean respectively of the input gradients wrt. randomly perturbed inputs. For our experiments, we use the squared version of SmoothGrad  \cite{hooker2019benchmark} as it outperformed its rooted counterpart. \citet{smilkov2017smoothgrad} also suggest an extension of Integrated Gradients, which merges Integrated Gradients with SmoothGrad, denoted by SmoothGrad-IG. Expected Gradients \cite{erion2019learning} is another derivative of Integrated Gradients that uses baselines drawn from the data distribution to aggregate values. 

The second class consists of modifications to the Vanilla Gradient method \cite{simonyan2013deep}, of which PatternAttribution is a member. Gradient times Input \cite{shrikumar2016not} simply multiplies the gradient saliency map by the input, while Guided Backpropagation \cite{springenberg2014striving} aims to filter for positive class evidence by inhibiting negative gradient values as well as those corresponding to negative values in the layer inputs. 
\section{Results \& Discussion}
Saliency maps produced by IG, PA and PGIG
are shown in Fig.~\ref{fig:heatmaps}. For comparability, we choose the same input image that \citet{kindermans2017learning} discuss in their paper. We observe that the heat map generated by PGIG appears plausible, as do the heatmaps of its parent methods: the most salient input features are located in the proximity of the African elephant in the input image. 

Regarding faithfulness, confidence curves are plotted in Fig.~\ref{fig:exp-pert-ig}. We observe that, according to the image degradation metric, the random patch ordering performs worst with VGG-16 on ImageNet, as expected. The random ordering is followed by the simple gradient method. It should be mentioned, however, that the simple gradient is more a sensitivity detector than an attribution method. 

Gradient times Input and Integrated Gradients both multiply gradients with inputs and perform similarly in our experiment. This is on par with the finding that in the linear case, Input times Gradient and Integrated Gradients even are equivalent methods \cite{adebayo2018sanity}. 

Both methods are surpassed by VarGrad, which itself is exceeded by SmoothGrad$^{2}$, SmoothGrad-IG, Expected Gradients, Guided Backpropagation and PatternAttribution; all of which perform similarly. Interestingly, these methods are of different classes: VarGrad, SmoothGrad, SmoothGrad-IG and Expected Gradients are gradient aggregating methods, whereas Guided Backpropagation and PatternAttribution are gradient modifying methods. Thus, we do not see any class membership preference.

Pattern-Guided Integrated Gradients is both, a gradient aggregating method and a gradient modifying method. According to the image degradation metric, it outperforms all other methods tested. However, this result is definitive only for methods without hyper parameters, such as Guided Backpropagation or PatternAttribution. We report the hyper parameters we use in the appendix.   
\section{Related Work}
Much of the related work that inspired the new method has already been mentioned in the previous sections. PGIG is of course based on its parent methods, IG \cite{sundararajan2017axiomatic} and PA \cite{kindermans2017learning}. PGIG is not the first method to extend IG, however. Expected Gradients \cite{erion2019learning} and a layered version of Intergrated Gradients \cite{mudrakarta2018did} are other examples. Integrated-Gradient Optimized Saliency \cite{qi2019visualizing} uses IG with mask optimization to generate attributions. Unlike PGIG, however, none of these methods apply informative directions.

 PGIG is both a modification to the Vanilla Gradient method \cite{simonyan2013deep}, such as Guided Backpropagation \cite{springenberg2014striving}, and a gradient aggregate method, such as SmoothGrad \cite{smilkov2017smoothgrad}, VarGrad \cite{adebayo2018sanity}, or the very recent SmoothTaylor method \cite{goh2020understanding}. For SmoothTaylor, \citet{goh2020understanding} bridge IG and SmoothGrad -- loosely related to what \citet{smilkov2017smoothgrad} propose for SmoothGrad-IG -- but within a Taylor's theorem framework. 
\section{Conclusion \& Future Work}
We present Pattern-Guided Integrated Gradients, which combines Integrated Gradients with PatternAttribution. Due to favorable properties, the new method passes stress tests that both parent methods fail. Furthermore, in a large-scale image degradation experiment, PGIG outperforms nine alternative methods, including its parent methods.

The image degradation metric that we offer to empirically validate the new method is being discussed, however \cite{hooker2019benchmark}. In the future, the new method should thus also be tested against other metrics. Furthermore, IG was shown to have a problematic degree of invariance to model randomization \cite{adebayo2018sanity}. It should be explored to what degree PGIG still exhibits this behaviour.    
\section*{Acknowledgements}
We would like to thank Christoph Alt, David Harbecke, Moritz Augustin and Arne Binder for their helpful feedback. Furthermore, we would like to thank Jonas Mikkelsen for helping with the code review. This work has been supported by the German Federal Ministry of Education and Research as part of the project BBDC2 (01IS18025E).
\bibliography{lit}
\bibliographystyle{icml2020}
\appendix
\section{Hyperparameters}
Below, we cite the formulas of the methods that involve hyper parameters and report the hyper parameters we use in our experiments. 

\subsection{(Pattern-Guided) Integrated Gradients}
Integrated Gradients \cite{sundararajan2017axiomatic} is given by
\begin{equation*}
\phi_{f,i}(x) = \frac{x_{i} - \bar{x}_{i}}{m}\sum_{k=1}^{m} \frac{\partial f(\bar{x} + \frac{k}{m}(x - \bar{x}))}{\partial x_{i}}
\end{equation*}

In our experiments, $\bar{x}=0, m=25$. For the proposed pattern-guided version we use the same hyperparameters.
\subsection{SmoothGrad$^{2}$}
SmoothGrad$^{2}$ \cite{hooker2019benchmark} is given by 
\begin{equation*}
    \phi_{f,i}(x) = \frac{1}{n}\sum_{1}^{n}\left(\frac{\partial f(x')}{\partial x'_{i}}\right)^{2}
\end{equation*}
where $x' = x + \mathcal{N}(\mu, \sigma^{2})$ is sampled for each $n$. In our experiments, $n=25, \mu=0, \sigma^{2}=0.15$.

\subsection{SmoothGrad-IG}
SmoothGrad-IG \cite{smilkov2017smoothgrad} is given by
\begin{equation*}
\phi_{f,i}(x) = \frac{x_{i} - \bar{x}_{i}}{m}
\sum_{k=1}^{m} 
\frac{1}{n}\sum_{1}^{n} \frac{\partial f(\bar{x} + \frac{k}{m}(x' - \bar{x})))}{\partial x'_{i}}
\end{equation*}
where $x'$ is defined above. In our experiments, $\bar{x}=0, m=25, n=25, \mu=0, \sigma^{2}=0.15$.

\subsection{VarGrad}
VarGrad \cite{hooker2019benchmark} is given by  
\begin{equation*}
    \phi_{f,i}(x) = var_{n}(\frac{\partial f(x')}{\partial x'_{i}})
\end{equation*}
where $x'$ is defined above. In our experiments, $n=25, \mu=0, \sigma^{2}=0.15$. 

\subsection{Expected Gradients}
Expected Gradients \cite{erion2019learning} is given by 
\begin{equation*}
    \phi_{f,i}(x) = \underset{\bar{x}\sim \mathbf{x}, \alpha \sim U(0,1)}{\mathbb{E}} \left[(x_{i} - \bar{x}_{i}) \frac{\partial f(\bar{x} + \alpha(x - \bar{x}))}{\partial x_{i}}\right]
\end{equation*}
In our experiments, we sampled $\bar{x}$ 49 times from the data. 

\end{document}